\documentclass[10pt,twocolumn,letterpaper]{article}

\usepackage{cvpr}              
\usepackage{float}
\usepackage{algorithm}
\usepackage{algorithmic}
\usepackage{xcolor}
\usepackage{url}

\definecolor{cvprblue}{rgb}{0.21,0.49,0.74}
\usepackage[pagebackref,breaklinks,colorlinks,allcolors=cvprblue]{hyperref}

\def\paperID{11000} 
\def\confName{CVPR}
\def\confYear{2025}

\title{Zero-to-Hero: Zero-Shot Initialization Empowering Reference-Based Video Appearance Editing}

\author{Tongtong Su$^{1,2}$, Chengyu Wang$^{2}$\footnotemark[1], Jun Huang$^2$, Dongming Lu$^{1}\footnotemark[1]$\\
$^1$ Zhejiang University,
$^2$ Alibaba Cloud Computing\\
{
\tt\small \{sutongtong,ldm\}@zju.edu.cn, 
    }\\
    {
\tt\small \{chengyu.wcy,huangjun.hj\}@alibaba-inc.com
    }
}

\begin{document}

\twocolumn[{
\maketitle
\begin{figure}[H]
\hsize=\textwidth 
\centering
\includegraphics[width=\textwidth]{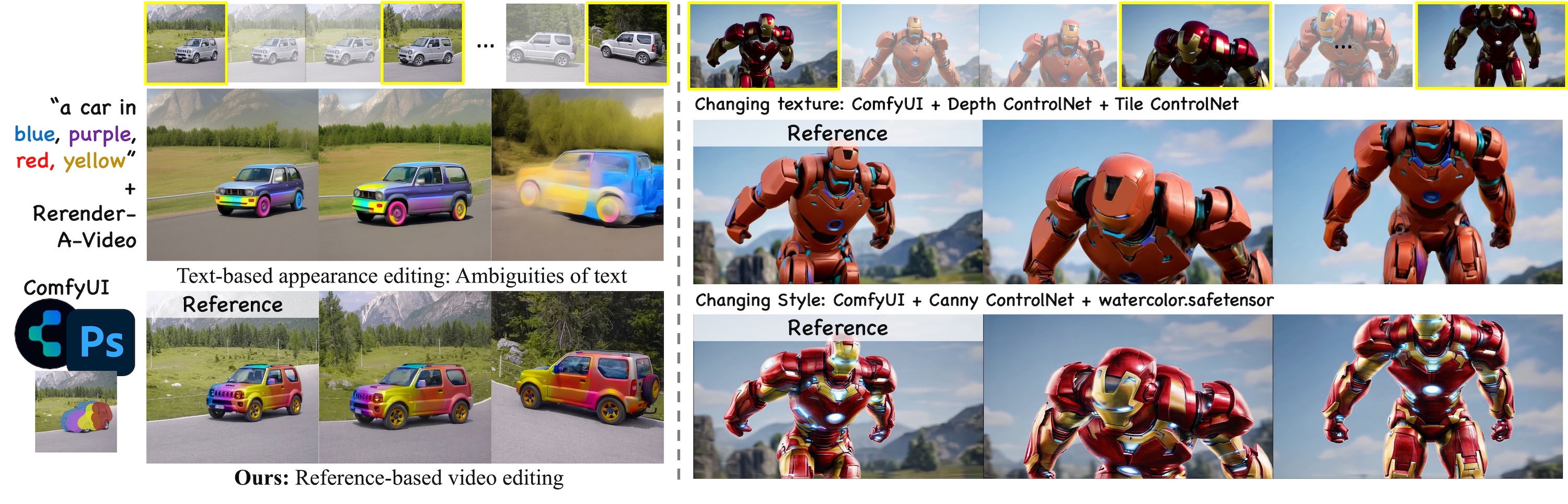}
\caption{\textbf{Left:} Our reference-based editing method enables users to precisely edit appearances by incorporating complex layouts of color with arbitrary tools such as Photoshop or ComfyUI to create references, then consistently propagate these edits to subsequent frames. \textbf{Right:} Our approach supports all spatially-aligned appearance editing, including texture and style.}
\label{fig:teaser}
\end{figure}
 }]

\renewcommand{\thefootnote}{\fnsymbol{footnote}}
\footnotetext[1]{Co-corresponding authors.}
\renewcommand{\thefootnote}{\arabic{footnote}}

\begin{abstract}

Appearance editing according to user needs is a pivotal task in video editing. Existing text-guided methods often lead to ambiguities regarding user intentions and restrict fine-grained control over editing specific aspects of objects. To overcome these limitations, this paper introduces a novel approach named {Zero-to-Hero}, which focuses on reference-based video editing that disentangles the editing process into two distinct problems. It achieves this by first editing an anchor frame to satisfy user requirements as a reference image and then consistently propagating its appearance across other frames. 
We leverage correspondence within the original frames to guide the attention mechanism, which is more robust than previously proposed optical flow or temporal modules in memory-friendly video generative models, especially when dealing with objects exhibiting large motions. It offers a solid \underline{ZERO}-shot initialization that ensures both accuracy and temporal consistency. However, intervention in the attention mechanism results in compounded imaging degradation with over-saturated colors and unknown blurring issues. Starting from Zero-Stage, our Hero-Stage \underline{H}olistically learns a conditional generative model for vid\underline{E}o \underline{R}est\underline{O}ration.
To accurately evaluate the consistency of the appearance, we construct a set of videos with multiple appearances using Blender, enabling a fine-grained and deterministic evaluation. Our method outperforms the best-performing baseline with a PSNR improvement of 2.6 dB. The project page is at~\url{https://github.com/Tonniia/Zero2Hero}.
\end{abstract}

\vspace{-1em}

\section{Introduction}


Video editing aims to modify the target video according to user demands. One of the most important sub-tasks is appearance editing~\cite{yang2023rerender,yang2024fresco,wang2024cove}, in which we preserve the structure of the target video frames while altering the color, texture of the object, or overall style. Previously, text-guided video editing addressed this task by leveraging pre-trained Text-to-Image (T2I) models, which rely on textual input (i.e., prompts) as the editing guidance signal~\cite{yang2023rerender,yang2024fresco,feng2024ccedit,geyer2023tokenflow,wang2024cove,cong2023flatten,kara2024rave}. However, ambiguities in text regarding user intentions may limit fine-grained control over the editing results. Therefore, a more practical solution for users to effectively convey their intentions is to explicitly provide a reference image, leading to the \emph{reference-based video editing} task~\cite{ku2024anyv2v,liu2024stablev2v,ouyang2024i2vedit}. It disentangles video editing into two problems: (1) editing a single image as a reference and (2) consistently propagating it to subsequent frames.



The first sub-task can be addressed using T2I models or arbitrary user manipulation through art design software, allowing for fine-grained appearance changing. The main difficulty lies in the second sub-task: \emph{how to consistently propagate the edited reference frame to other frames}. Current propagation methods can be divided into two groups. The first group of methods uses optical flow obtained from the target video to guide the propagation of reference image features~\cite{yang2024fresco,yang2023rerender,cong2023flatten}. The performance of these methods can be limited by optical flow estimation~\cite{xu2022gmflow}, which was trained on a specific set of videos. Consequently, its accuracy noticeably degrades when dealing with videos involving significant motion. The second group~\cite{ku2024anyv2v,liu2024stablev2v,ouyang2024i2vedit} leverages Image-to-Video (I2V) models~\cite{zhang2023i2vgen,SVD} to invert the target video to noisy latent representations, then uses the reference image as a guidance signal to denoise. However, the video length is constrained by the memory demands of inversion, and the temporal modeling limitations of these memory-friendly I2V models also restrict the range of motion. Recent work~\cite{ouyang2024i2vedit} fine-tunes the I2V model with specific target videos. However, for videos with significant motion that deviate far from the I2V domain, it is challenging to strike a balance between adequately fitting the motion pattern and preventing overfitting its appearance. 




In this work, we explore propagation-based methods but redefine the propagation problem as the more general appearance transfer task~\cite{tumanyan2022splicing,park2020swapping,mou2023dragondiffusion}: maintaining the structure of the target image while utilizing the appearance of the reference image. This task involves finding the correspondence between reference and target images and then propagating the reference image features into the target ones. Recent approaches connect this task with the self-attention (SA) mechanism in diffusion models~\cite{mou2024diffeditor,mou2023dragondiffusion,epstein2023diffusion}, leveraging their generative capabilities to support zero-shot appearance transfer. Diffusion models can inherently model intra-similarity for correspondence and simultaneously propagate features using SA. Given two images, expanding SA to cross-image attention (CiA) is a common method for fusing features between images~\cite{chung2024style,alaluf2024cross,tewel2024training}. However, basic CiA can only capture coarse-grained correspondence, as the query of the target image exhibits similarity to many keys in the reference image~\cite{alaluf2024cross}. The weighted averaging of matched values leads to a loss of fine-grained details and limits the ability to handle fine-grained appearance transfer. Direct applying contrast value to the attention map can introduce inaccurate transfer since the attention map at the early denoising stage, with a high noise level, cannot represent accurate correspondence. Some research~\cite{tang2023emergent,luo2024diffusion,zhang2024tale} has found that DIffusion FeaTures (DIFT) at certain timesteps and U-Net layers can best represent correspondence.


Obtaining correspondence is merely the first step. Directly performing pixel-level swapping based on the highest similarity without introducing a generative process remains highly sensitive to occasional inaccurate matching, often leading to artifacts, prominently marked by noticeable patch splitting~\cite{zhang2024tale}. Using correspondence to guide CiA during denoising is more robust, given the output domain constraint of the generative model. This is implemented through an attention mask, i.e., setting the top-$k$ entries in correspondence to 1 to construct the mask. When $k=h\times w$, it is the basic CiA, facing the same problem of weighted averaging of values from the reference. With gradually decreasing $k$, the transfer becomes more accurate. However, this also progressively over-intervenes in the attention maps, leading to a compounded degradation of over-saturated colors and blur patterns of unknown origin~\cite{ahn2024self,liu2025understanding}. There exists an upper limit to what can be achieved with zero-shot methods.


In this paper, we propose \emph{Zero-to-Hero}, which builds upon the aforementioned \underline{ZERO}-shot intermediate result (i.e., Zero-Stage) as a good initialization and incorporates the Hero-Stage for \underline{H}olistic vid\underline{E}o \underline{R}est\underline{O}ration. This approach can be formulated as a conditional generation problem~\cite{zhang2023adding}, where the training data pairs require ground truth. The sole ground truth available is the reference on the anchor frame. We observed that the Zero-Stage exhibits a similar pattern of degradation across all frames, indicating that training on the anchor frame has the potential to generalize effectively to all subsequent frames. To accelerate training convergence, the original frame is utilized as an auxiliary condition to encourage shortcut mapping for non-edited regions. To evaluate appearance consistency more accurately beyond semantic-level CLIP-based scores~\cite{huang2024vbench}, we collect a set of 3D objects with multiple appearances and render them under significant camera motion to construct videos using \emph{Blender}. This supports fine-grained and deterministic evaluation. Our method outperforms the best-performing baseline with a PSNR improvement of 2.6 dB.

\section{Related Work}
\textbf{Reference-based Video Appearance Editing.} Text-guided video editing addresses this task by leveraging pre-trained Text-to-Image (T2I) models, which rely on textual input (i.e., prompts) as the editing guidance signal~\cite{yang2023rerender,yang2024fresco,geyer2023tokenflow,wang2024cove}. However, ambiguities in text regarding user intentions may limit fine-grained control over the editing process. Therefore, a more practical solution for users to effectively express their intentions is to use a single image, leading to \emph{reference-based video editing}~\cite{ku2024anyv2v,liu2024stablev2v,ouyang2024i2vedit}, which offers a more flexible solution for video editing. AnyV2V~\cite{ku2024anyv2v} adopts the earlier memory-friendly I2V model, I2VGen~\cite{zhang2023i2vgen}, for DDIM inversion~\cite{song2020denoising}. During the denoising steps, intermediate features are selectively injected to preserve the original motion. This process requires fine-grained hyperparameter tuning of different injection rates at both spatial and temporal modules. Instead of relying on the original temporal module, I2VEdit~\cite{liu2024stablev2v} fine-tunes it to fit the specific target video.


One related reference-based video processing task is video colorization~\cite{yang2024bistnet,yang2024colormnet}. The colored image retains the exact same grayscale as the uncolored version. It cannot transfer a light color to a dark region within a grayscale image, nor change the texture or style for general editing tasks~(as illustrated in Figure~\ref{fig:ablation_mode}).

\noindent\textbf{Spatial-Aligned Conditional Generative Model.} Spatial-aligned conditional generation~\cite{zhang2023adding,mou2024t2i,qin2023unicontrol} has been extensively studied in the context of diffusion models. Notable methods include ControlNet~\cite{zhang2023adding}, which introduces a residual branch to process the conditional image and applies spatially-aligned additions to the main branch. Data collection involves collecting large amounts of images and applies the corresponding image processing techniques~(e.g., Canny edge detection or depth estimation) to generate conditions, forming training pairs. This can be regarded as obtaining shared-pattern degradation. Various common types of known degradation, such as Gaussian Blur~\cite{blur-controlnet}, Tile~\cite{tile-controlnet}, and grayscale~\cite{color-controlnet}, have corresponding ControlNet. In our problem setting, correspondence as CiA guidance can be regarded as a compounded degradation. Recently, DiT-based diffusion models~\cite{chen2023pixart,peebles2023scalable,flux} have demonstrated significant advantages over U-Net models in terms of imaging quality and prompt understanding. Many works~\cite{tan2024ominicontrol,zhang2025easycontrol,tan2025ominicontrol2} fine-tuning these base model for conditional generation, demonstrating faster convergence and reduced training data requirements compared to ControlNet.


\section{Method}

Given a set of consistent sequences of images from a specific video, we select one anchor frame $I^{\text{anc}}$ and edit it exclusively at the appearance level (ensuring spatial alignment, e.g., using ComfyUI with Canny ControlNet) to obtain the reference frame $I^{\text{ref}}$. For each target frame $I^{\text{tgt}}$ of the output video, we compose a triplet: $(I^{\text{anc}}, I^{\text{ref}}, I^{\text{tgt}})$. Our goal is to generate the output image frame $I^{\text{out}}$, which depicts the structure present in $I^{\text{tgt}}$ while incorporating the appearance edited in $I^{\text{ref}}$. The frame $I^{\text{anc}}$ serves as a connection since $I^{\text{anc}}$ and $I^{\text{ref}}$ are spatially aligned, and the matching between $I^{\text{anc}}$ and $I^{\text{tgt}}$ is termed correspondence~\cite{zhang2024tale,luo2024diffusion,tang2023emergent}. In our work, we utilize a pre-trained Stable Diffusion model~\cite{LDM}, with VAE encoding the image $I$ into the latent representation $z_0$, and DDIM inversion~\cite{song2020denoising} to obtain the noisy latent $z_t$. During inversion, attention features in the intermediate steps are preserved. Similar to previous works~\cite{alaluf2024cross,chung2024style}, our method produces an image from a denoising process starting from $z_{t}^{\text{tgt}}$, with feature injections from the reference image. This process is referred to as Cross-image Attention, which is an extension of Self-Attention. We first review these two mechanisms.

\begin{figure*}[h]
  \centering
  \includegraphics[width=\linewidth]{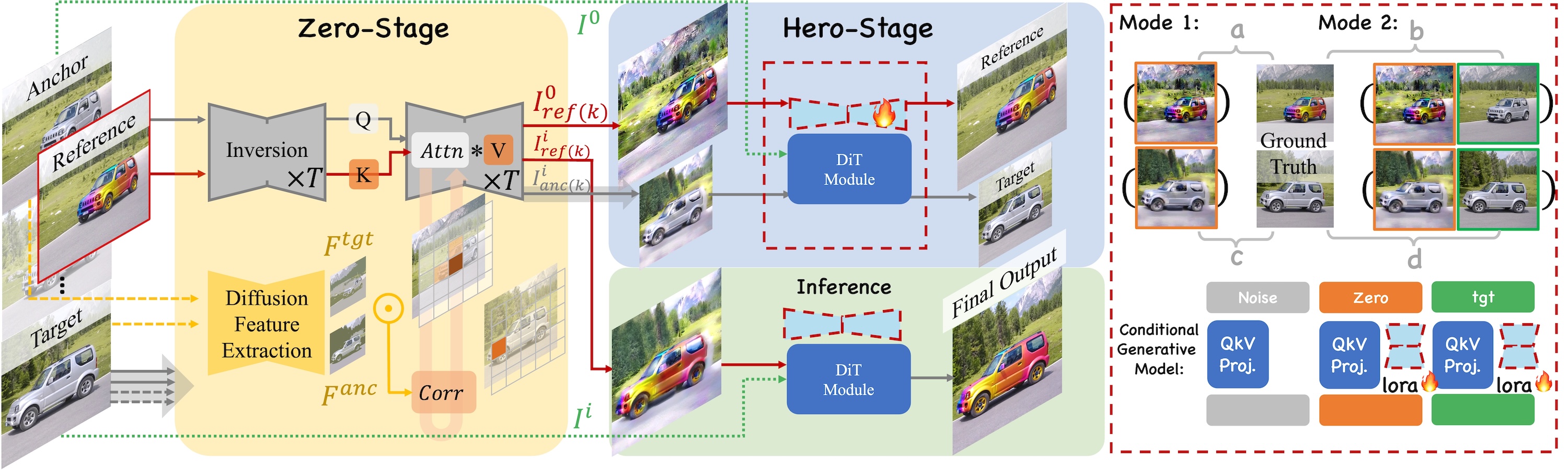}
\caption{Our framework. \textbf{Zero-Stage:} Correspondences ($Corr$) estimated from the anchor and target frames are utilized to guide Cross-image Attention ($Attn$) between the reference and anchor frames, enabling accurate appearance transfer in a zero-shot manner. \textbf{Hero-Stage:} We learn a conditional generative model by incorporating LoRA to process conditional tokens. There are two modes of condition injection: one condition with one LoRA (Mode 1) and two conditions with two independent LoRAs (Mode 2). Four pairs of images serve as potential training data, from a to d (see Table~\ref{tab:data_pairs}).}
  
\end{figure*}

\subsection{Preliminaries}
\noindent\textbf{Self-Attention~(SA) and Cross-image Attention~(CiA).} Self-Attention (SA) serves as a fundamental component in diffusion models for establishing the global structure. Given an input latent $z_{t}$ comprising $h \times w$ tokens, the intermediate feature in the U-Net $\phi(z_{t})$ employs SA linear projections $\ell_{q}$, $\ell_{k}$, and $\ell_{v}$, which map the input image features $z_{t}$ onto the query, key, and value matrices of a specified dimension $d$: 
$Q = \ell_{q}(\phi(z_{t})), \quad K = \ell_{k}(\phi(z_{t})), \quad V = \ell_{v}(\phi(z_{t}))$,
respectively. The attention map $Attn$ is defined as:
$Attn = \text{Softmax}\left(\frac{Q \cdot K^{\mathsf{T}}}{\sqrt{d}}\right)$,
which computes the similarity among the tokens. The output is defined as the aggregated feature of $V$ weighted by similarity, denoted as $\phi(z_{t}) = Attn \cdot V$. $Attn$ can represent the structure of a target image when applying DDIM inversion~\cite{song2020denoising}, while $V$ contains appearance information. During inversion, intermediate $Q^{\text{tgt}}, K^{\text{tgt}}, V^{\text{tgt}}$ are saved and selected for injection into the denoising process for different tasks, e.g., editing~\cite{tumanyan2023plug,hertz2022prompt}, style transfer~\cite{chung2024style,xu2024freetuner}.

\begin{figure}[h]
  \centering
  \includegraphics[width=\linewidth]{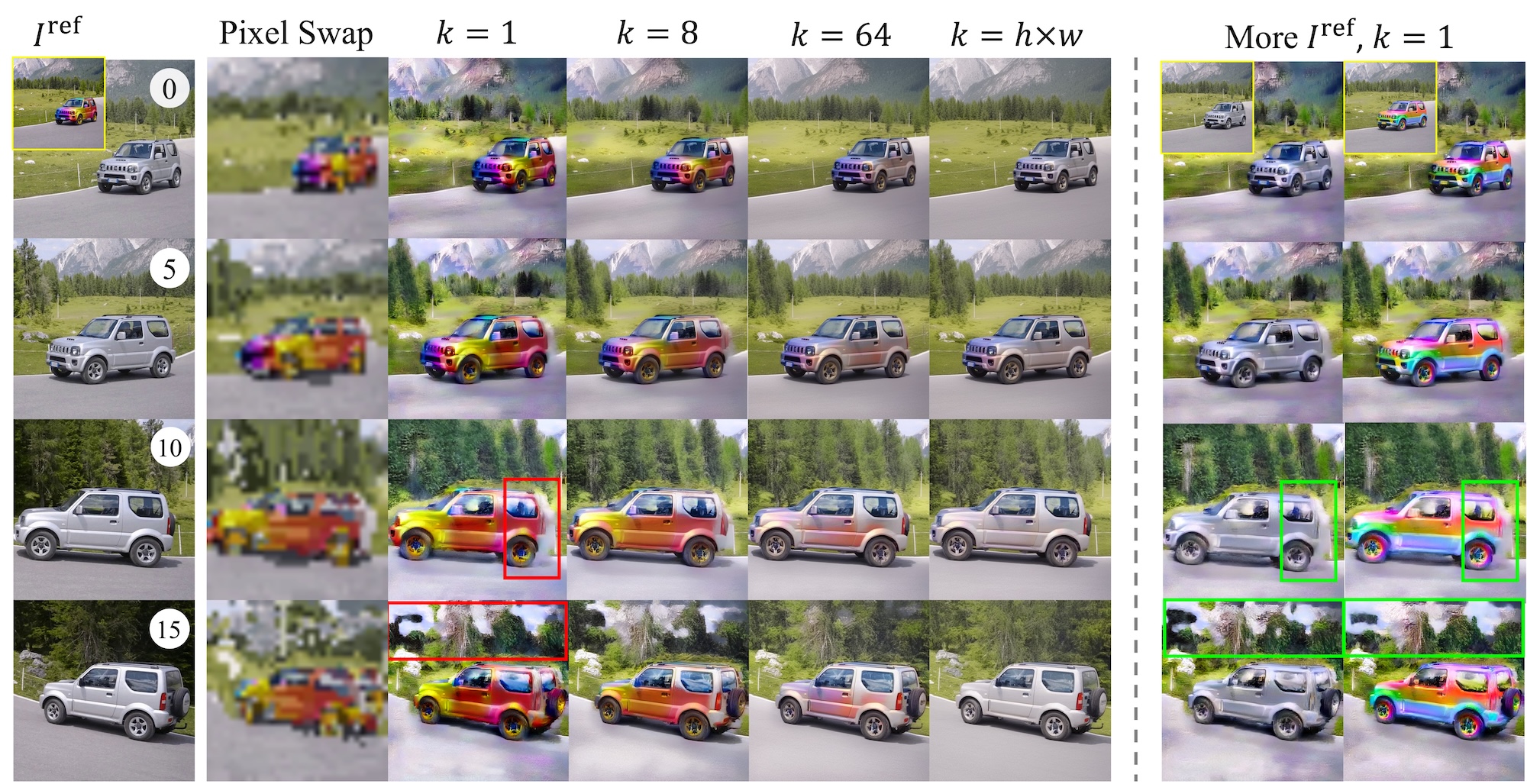}
\caption{\textbf{Left:} $Corr$ guidance with increasing $k$. When $k=h\times w$, it     corresponds to the original Cross-image Attention. \textbf{Right:} Using other references results in similar missing patterns (red and green boxes).}
  
  \label{fig:corr_k}
\end{figure}

Cross-image Attention (CiA) extends the concept of Self-Attention (SA) to multiple images. When $Q$ is derived from the target image, and $K$ and $V$ come from a reference image, CiA measures the similarity between tokens from the target (tgt) and reference (ref) images:
$Attn = \text{Softmax}\left(\frac{Q^{\text{tgt}} \cdot {K^{\text{ref}}}^{\mathsf{T}}}{\sqrt{d}}\right)$.
This similarity weights the reference $V^{\text{ref}}$ to transfer information to the target output: $Attn \cdot V^{\text{ref}}$, which is used in style transfer tasks~\cite{chung2024style}. $K^{\text{ref}}$ and $V^{\text{ref}}$ can be extended to multiple images, which is beneficial in video processing tasks~\cite{qi2023fatezero,yang2023rerender}. Although CiA represents similarity and is useful for style transfer, it does not ensure accurate correspondence between $I^{\text{ref}}$ and $I^{\text{tgt}}$. The distribution of CiA is scattered, implying that a token in $I^{\text{tgt}}$ may interact with numerous tokens in $I^{\text{ref}}$. This interaction averages their $V^{\text{ref}}$, including irrelevant ones, and ultimately leads to the color leakage problem (as shown in the last column of Figure~\ref{fig:corr_k}). Some works introduce a temperature $\tau$ to enhance the contrast of attention maps, encouraging focus on a few patches~\cite{chung2024style}. Others boost contrast by increasing the variance of the attention maps~\cite{alaluf2024cross}. However, CiA still struggles to establish correspondence for spatially unaligned samples in videos with significant motion, making accurate appearance transfer a research challenge.


\noindent\textbf{Correspondence from Diffusion Features.} Diffusion models exhibit strong semantic feature extraction capabilities~\cite{zhang2024tale,luo2024diffusion,tang2023emergent}. These studies investigate which intermediate DIffusion FeaTures (DIFT) are the most effective for establishing semantic correspondence. They add noise at a specific timestep $t$ and feed the noisy latent into the U-Net. Intermediate features from the decoder are extracted through a single denoising step. In our work, we denote intermediate features as $F$. Similarly to $Attn$ in CiA, the semantic correspondence is based on dot product similarity: $Corr = F^\text{tgt} \cdot {F^\text{anc}}^{\mathsf{T}}$. The correspondence between $F^\text{tgt}$ and $F^\text{anc}$ is more accurate than that between $F^\text{tgt}$ and $F^\text{ref}$, as both are derived from the original video. Since $F^\text{anc}$ and $F^\text{ref}$ are spatially aligned, $Corr$ can be used to guide CiA between $F^\text{tgt}$ and $F^\text{ref}$.


Previous works have exploited this correspondence for one-to-one pixel-level swapping~\cite{zhang2024tale}. The output often contains noticeable pixel boundaries and artifacts caused by inaccurate matching. A recent study~\cite{wang2024cove} adapts a sliding-window-based strategy based on the target video. However, as a text-guided method, it heavily relies on DDIM inversion of the original frames and text prompts, without explicitly incorporating $V$ from the reference. This significantly limits its ability to perform free editing with complex appearance changes. Additionally, it requires many hyperparameters for each case, such as the number of steps for DDIM feature injection and the size of the sliding window. This is an inherent drawback of training-free methods.





\subsection{Zero-Stage: Correspondence as CiA Guidance}

We propose using correspondence as accurate guidance for CiA $Attn$, through the implementation of masked attention. The mask is created by selecting the top-$k$ entries in $Corr$ and setting these positions to 1 in the mask matrix $M(Corr, k)$, where $k$ ranges from 1 to the total number of tokens $h \times w$. These selected places are assigned a large value in the original attention score matrix $A$, resulting in $A \oplus M(Corr, k)$, before the softmax operation. Using this guided attention and $V^\text{ref}$ from the reference, a denoising step is:
\begin{equation}
\hat{z}_{t-1}^\text{tgt} = \epsilon_\theta(z_t^\text{tgt}, A \oplus M(Corr, k), V^\text{ref})
\end{equation}
where $\hat{z}_{t-1}^\text{tgt}$ will be iteratively denoised until obtaining the clean latent $\hat{z}_{0}^\text{tgt}$, thereby achieving appearance transfer. The final transferred result of the target frame $I^\text{tgt}$ is termed as $I_{M(k)}^\text{tgt}$.

The selection of $k$ is challenging. As shown in Figure~\ref{fig:corr_k}, we apply different $k$ values on four frames, including anchor frame 0. When $k=1$, the accurate semantic correspondence can successfully transfer each part of the car to the target image with corresponding colors. However, the modifications applied to the original attention mechanism are substantial, leading to compounded degradation in imaging quality, including irregular blurring and color over-saturation. The former is caused by the collapsed structure within the attention map, while the latter arises from the increased feature magnitude due to the low entropy of the identity matrix~\cite{liu2025understanding}. As $k$ increases, the imaging quality gradually returns to normal, while the influence of $Corr$ guidance decreases. When $k=h\times w$, it is equivalent to the original CiA, facing the same problems of color leakage and inaccurate appearance transfer.

Based on the above analysis, it is difficult to find an appropriate $k$ that achieves a satisfying result across the whole image in zero-shot manner. Intuitively, imaging degradation is less harmful than color leakage and offers greater potential for post-processing. We observed that, for different frames, the degradation pattern is similar. We also have the ground-truth high-quality version of $I_{M(k)}^\text{anc}$, which is $I^\text{ref}$. Therefore, a natural question arises: \textit{can we train a conditional generative model to restore images with compounded degradation caused by masked attention?}



\subsection{Hero-Stage: Zero-Stage as Condition}

We use the intermediate result from the Zero-Stage as the image condition and perform \underline{H}olistic vid\underline{E}o \underline{R}est\underline{O}ration, with irregular blurring and color over-saturation issues. When training on the pair $\{I_{M(k)}^\text{anc}, I^\text{ref}\}$, it certainly guarantees fitting on this anchor frame. We investigate its ability to generalize to subsequent frames, particularly long-range frames with substantial motion.



\subsubsection{Generalization Across Video}



For target frames that are near the anchor frame (e.g., frame 5 in Figure~\ref{fig:corr_k}), the color over-saturation and blurring patterns are similar to the anchor frame. However, for long-range target frames (frames 10 and 15), in addition to the two aforementioned aspects of degradation, there is also a noticeable color-missing issue caused by unsuccessful matching in significantly changed backgrounds (highlighted in the red box). This color-missing issue does not occur in the training anchor frame pair, so the trained model is more likely to preserve these missing parts instead of restoring them. Extracting masks for unchanged regions and straightforward replacement of the original target can be sensitive to mask extraction errors. These methods may also fail due to the lack of clear segmentation between objects and the background. To address this, we aim to incorporate the target frames into the training pipeline, automatically providing auxiliary information. There are two potential ways to use target frames $I^\text{tgt}$:


\noindent\textbf{Mode 1: Implicit Usage}. We encourage the model to implicitly learn unseen regions through the provision of auxiliary training pairs. We use $I^\text{anc}$ as the reference to construct a series of frames $\{I^{0}_{\text{anc}(k)},\cdots,I^{n}_{\text{anc}(k)}\}$. As shown in Figure~\ref{fig:corr_k} on the right, frames 10 and 15, with other references, display similar missing regions (in the green box) as when $I^\text{ref}$ is used as the reference. Specifically, when using $I^\text{anc}$ as the reference, the corresponding ground truth for each target frame should be to reconstruct themselves. These auxiliary training pairs are termed zero-tgt: $\{I_{\text{anc}(k)}^i, I^{i}\}, i=[0,n]$.



\noindent\textbf{Mode 2: Explicit Usage}. We can also explicitly use original target frames by introducing an additional conditional branch, denoted as $\{(I_{\text{ref}(k)}^0, I^0), I^\text{ref}\}$. Ideally, $I_{\text{ref}(k)}^0$ provides an appearance-transferred intermediate result, while $I^0$ provides a shortcut to quickly reconstruct the unchanged region. This training pair is termed (zero, tgt)-ref. Similarly, we can construct (zero, tgt)-tgt pairs: $\{(I_{\text{anc}(k)}^i, I^i), I^i\}$.

In summary, all potential training pairs are shown in Table~\ref{tab:data_pairs}. Mode 1 contains a + c, and Mode 2 contains b + d. Regardless of the usage, fitting the anchor frame can always be guaranteed. The key challenge lies in ensuring that the target frame provides only auxiliary information of the unchanged region while preventing the leakage of its appearance.

\begin{table}[t]
    \centering
    \begin{tabular}{c|c|c}
    \toprule
        a. zero-ref & $\{I_{\text{ref}(k)}^0, I^\text{ref}\}$ & Mode 1 \\
        b. (zero, tgt)-ref & $\{(I_{\text{ref}(k)}^0, I^0), I^\text{ref}\}$ & Mode 2 \\
    \midrule
        c. zero-tgt & $\{I_{\text{anc}(k)}^i, I^i\}$ & Mode 1 \\
        d. (zero, tgt)-tgt & $\{(I_{\text{anc}(k)}^i, I^i), I^i\}$ & - \\
    \bottomrule
    \end{tabular}
    \caption{Pairs of training data. Mode 1 use a and c, Mode 2 use b. d may cause the model to overfit on the target $I^i$ due to the shortcut.}
    \label{tab:data_pairs}
    \vspace{-1em}
\end{table}

\subsubsection{Explicit Condition Avoids Leakage}

\begin{figure}[t]
  \centering
  \includegraphics[width=0.95\linewidth]{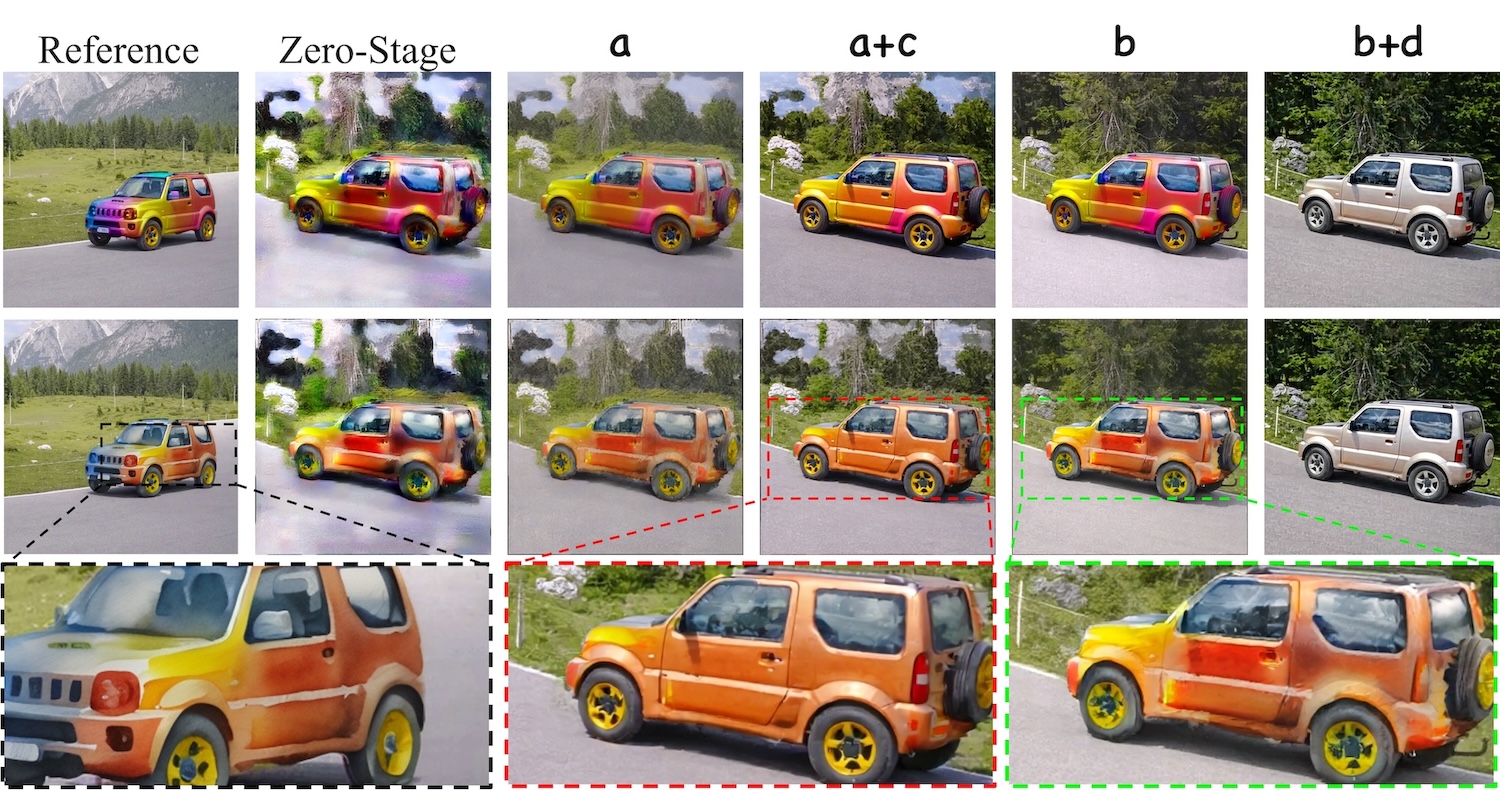}
  \caption{Mode 1~(a+c) can better preserve target structure of car than only using a, but it struggles with style transfer (e.g., watercolor in the second row) and restoring severely missing background regions. Mode 2~(b) can solve two problems. d will result in appearance leakage of target frame.}
  \label{fig:conditional_mode}
\end{figure}

As shown in Figure~\ref{fig:conditional_mode}, Mode 1 with a + c achieves better preservation of the outline of the car compared to only a, although the missing regions in the background remain unrepairable. In Mode 2, with the explicit injection of the target frame, the background can be successfully repaired. This suggests that the explicit mode can differentiate between edited and non-edited regions: for the edited object regions, it restores the Zero-Stage intermediate results, while for unchanged regions, it directly uses a shortcut to copy from the target frame.

When the editing type involves texture or style rather than only color change (second row, reference is watercolor style), Mode 1 with training pair c forces the model to learn zero-tgt on each target frame, causing the model to overfit them. As a result, the style is not successfully changed, and the color layout from the Zero-Stage is not used (zoom in on the red box). Conversely, Mode 2 explicit injection appears likely to leak the appearance of the target image, but it does not. This indicates that the training of the two conditional branches does not cause conflicts, and they can clearly differentiate their respective roles. The appearance of the subsequent frames is successfully transferred to the watercolor style (zoom in on the green box). For data pair d, $\{(I_{\text{anc}(k)}^i, I^i), I^i\}$, the model identifies shortcuts from the second condition to the target frame $I^i$. Even if the first condition is changed to $\{I_{\text{ref}(k)}^i\}$, the output still directly corresponds to the second condition. Therefore, it should not be involved in training.

\begin{table*}[h]
    \centering
    \resizebox{0.95\textwidth}{!}
    {\begin{tabular}{c|ccc|ccc|ccccc}
    \toprule
      & \multicolumn{3}{c}{Colorization} & \multicolumn{3}{c}{Blender-Color-Edit} & \multicolumn{5}{c}{General-Edit} \\
     \midrule
     Method & PSNR~($\uparrow$) &  LPIPS~($\downarrow$)  &  SSIM~($\uparrow$) & PSNR~($\uparrow$) &  LPIPS~($\downarrow$)  &  SSIM~($\uparrow$) &
     MS~($\uparrow$) & SC~($\uparrow$) & $\text{PSNR}_{\dagger}$~($\uparrow$) &  $\text{LPIPS}_{\dagger}$~($\downarrow$)  &  $\text{SSIM}_{\dagger}$~($\uparrow$) \\
     \midrule
    AnyV2V   & 22.7450  & 0.1456 & 0.7703 & 23.3208 & 0.1316 & 0.8174 & 0.9192 & 0.8325 & 26.2379 & 0.0935 & 0.8392 \\
    I2VEdit              & 23.4085 & 0.1231 & 0.8219 & 24.1103 & 0.1317 & 0.8044 & 0.9329 & 0.8724 & \textbf{27.0925} & \underline{0.0830} & \underline{0.8639} \\

    CiA~($\beta^*$)   & 23.2932 & 0.1486 & 0.7975 & 23.0619 & 0.1377 & 0.7686 & \underline{0.9409} & \textbf{0.9158} & 25.5291 & 0.0943 & 0.8239 \\
    
    Ours    & \textbf{28.2063 }& \textbf{0.0491} & \textbf{0.9298} & \textbf{26.7640} & \textbf{0.0565} & \textbf{0.8546} & \textbf{0.9428} & \underline{0.8978} & \underline{26.7768} & \textbf{0.0558} & \textbf{0.8886} \\
    \bottomrule
    \end{tabular}
   }
    \caption{Results on three appearance editing tasks for all reference-guided methods. For Colorization and Blender-Color-Edit, all frames have ground truth to calculate PSNR (dB), LPIPS, and SSIM. For General-Edit, calculations can only be performed on the anchor frame ($\dagger$) using the reference as ground truth. Motion Smoothness (MS) and Subject Consistency (SC) are utilized to evaluate temporal consistency.}
    \label{tab:baselines}
\end{table*}

\subsubsection{Balancing Condition Branches}

Since we have the Hero-Stage, can we use less-than-perfect intermediate results in the Zero-Stage, such as performing direct pixel swapping without introducing a generative process, to save time? As shown in Figure~\ref{fig:corr_k}, pixel swapping according to correspondence can also provide an initialization of the transferred result, although it contains less information or might be inaccurate. Can it achieve the same level of conditional control as our Zero-Stage initialization?


We emphasize the importance of \emph{condition balance}, i.e., two branches should ideally contain equal amounts of information. Otherwise, the model will be overly reliant on the side that provides more information. Since pixel swapping does not provide enough structural information but only the color layout, it will force the model to rely too much on the second condition, the target frame, to acquire structure and details of texture for the reconstruction of the reference. When the target frame has a totally different texture or style compared to the reference, the model will be confused about condition usage. In the most likely scenario, this degenerates into a recolor task, i.e., using pixel swapping to recolor the target image.

For the implementation of conditional generative models, in the past UNet-diffusion era, it was always implemented by ControlNet~\cite{zhang2023adding}, with high computational cost and slow convergence. In our task, we only need to focus on a set of images from one video, and the conditional degraded image itself contains sufficient appearance information, which is much simpler than training a general conditional model like canny or depth. Recently, with the development of DiT-based diffusion models, injection paradigms have largely changed. These methods~\cite{tan2024ominicontrol,zhang2025easycontrol} typically convert conditional images into image token sequences, concatenate them with noise latent representations, and fine-tune DiT modules with LoRA to process conditional tokens. The conditional tokens share the same positional encoding as the corresponding spatially-aligned noisy tokens, which accelerates convergence. With this efficient fine-tuning architecture and our solid Zero-Stage initialization, our Hero-Stage achieves faster convergence. For mode 2, two independent LoRAs are necessary for model to distinguish different control purpose.

\section{Experiments}
\subsection{Experimental Settings}
\subsubsection{Datasets} 
\begin{figure}[h]
  \centering
  \includegraphics[width=0.8\linewidth]{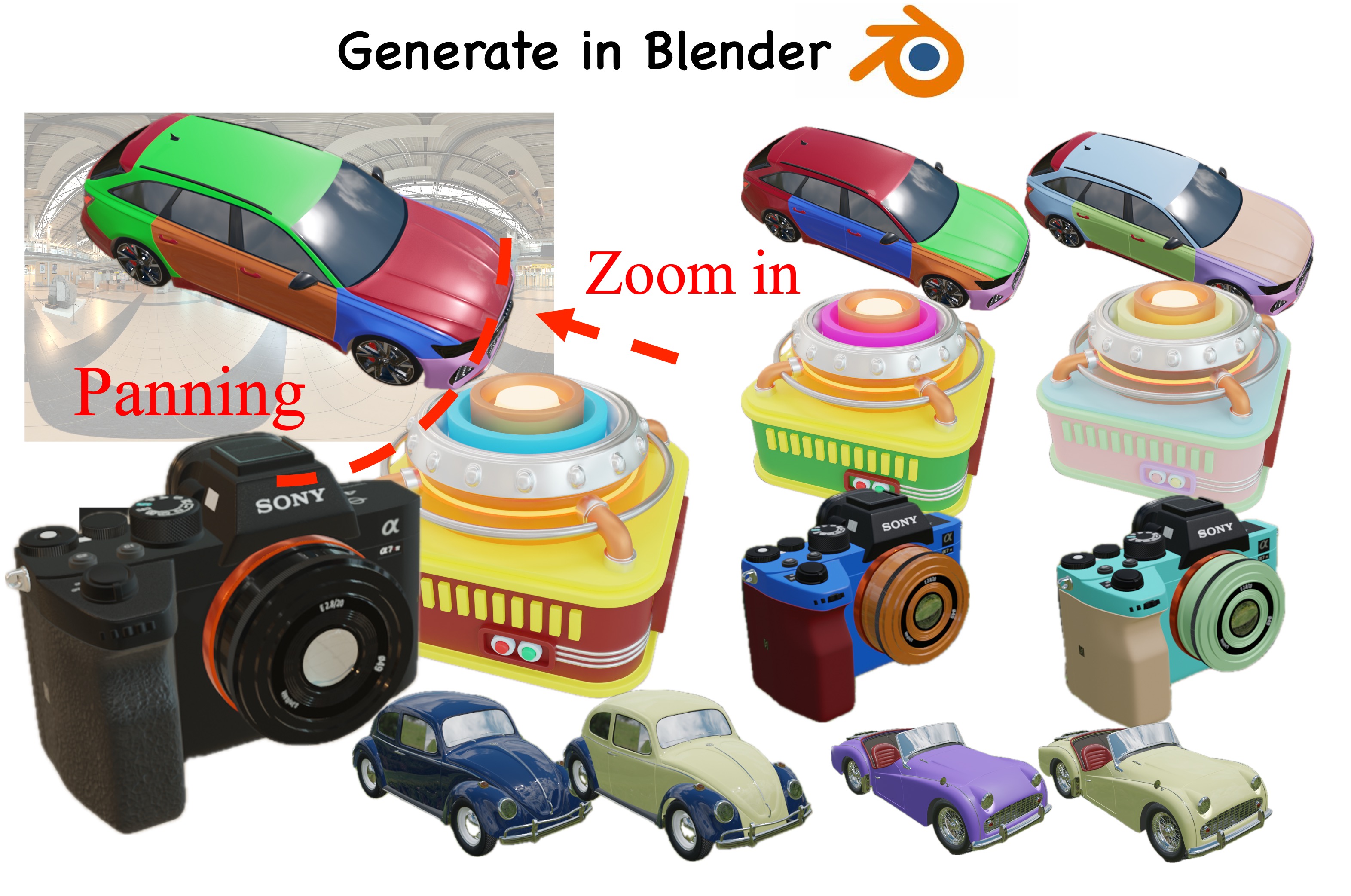}
  \caption{Blender-Color-Edit dataset, rendered in Blender.}
  \label{fig:blender_dataset}
\end{figure}

We utilize datasets from two sources. We use a collection of videos utilized in video editing baselines~\cite{yang2024fresco}. These videos have all been uniformly sampled to 16 frames. For each video, we experiment on two sub-tasks. The first sub-task is Colorization, where we convert the target video to grayscale, and the reference is the original anchor frame. The second sub-task is general appearance editing~(General-Edit), where the references are processed using Stable Diffusion WebUI with multiple ControlNets for spatial alignment. The editing types include color changes~(\emph{a car in the color red, blue, etc.}), texture changes~(\emph{a car made of clay}), and style changes~(\emph{a car in watercolor style}). The colorization task supports deterministic evaluation metrics including PSNR, LPIPS, and SSIM for each frame. For General-Edit, there is no ground truth available except for the anchor frame. Therefore, we employ auxiliary metrics to evaluate temporal consistency, including Motion Smoothness (MS) and Subject Consistency (SC), as proposed in VBench~\cite{huang2024vbench}. The colorization task does not cover the general appearance editing task, and General-Edit lacks ground truth, making the evaluation indirect. To address this, we construct a dataset using Blender. We collect five 3D objects, each prepared with two distinct appearances. We employ two classic camera movements: panning and zooming in. The dataset, Blender-Color-Edit, is illustrated in Figure~\ref{fig:blender_dataset}.

\begin{figure*}[h]
  \centering
  \includegraphics[width=\linewidth]{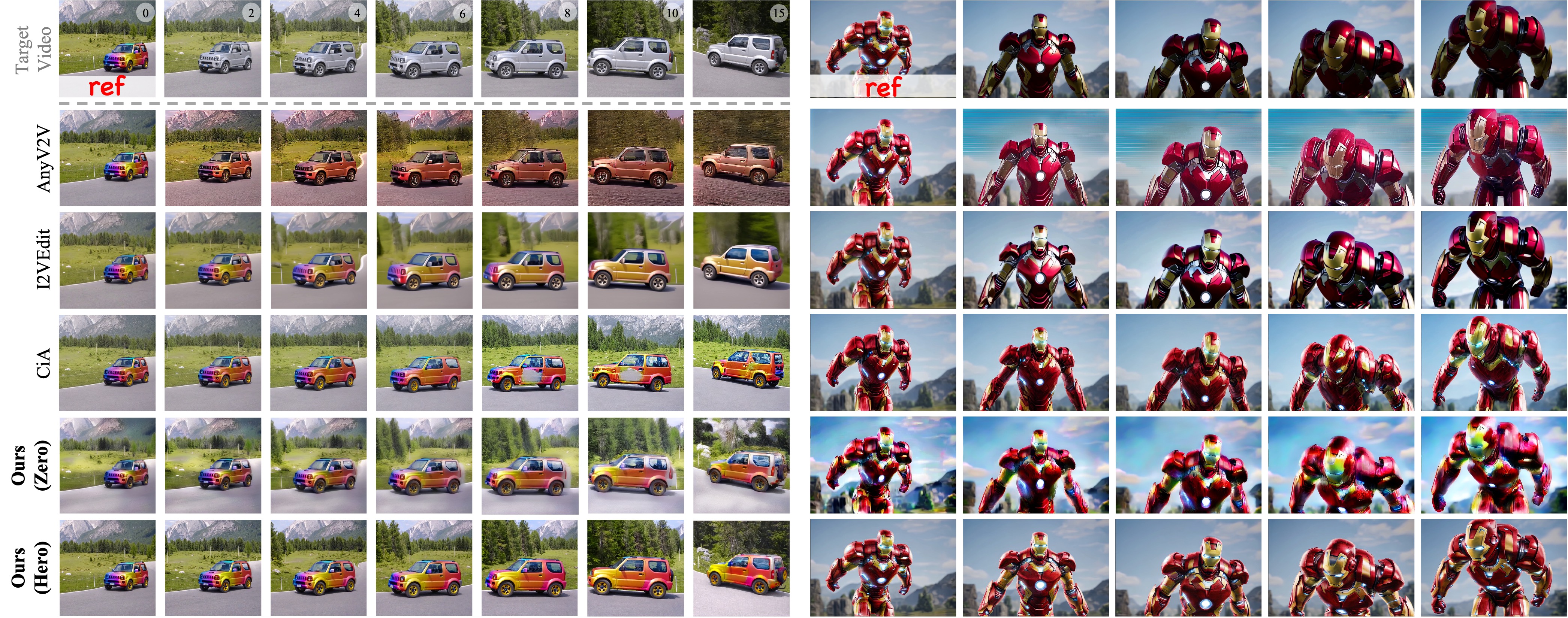}
  \caption{Qualitative results on General-Edit dataset. Our method maintains the highest consistent fidelity to reference appearance and target video structure.}
  \label{fig:baselines}
\end{figure*}

\subsubsection{Baselines}

We compare our approach with two reference-based video editing methods, namely AnyV2V~\cite{ku2024anyv2v} and I2VEdit~\cite{ouyang2024i2vedit}, along with an appearance transfer method, CiA with attention contrast~\cite{alaluf2024cross} employing Stable Diffusion. AnyV2V directly utilizes I2VGen-XL, applying DDIM Inversion and selectively injecting features during the denoising process. It requires varying hyperparameters for each scenario, including the injection rate at spatial, temporal, and feedforward modules. Following its setting, we adopt an inversion step of $500$ and a denoising step of $50$, traversing the rates across $[0.2, 0.5, 0.8]$, and selecting the optimal result for each scenario. 
I2VEdit initially performs LoRA fine-tuning of the temporal module of SVD to adapt to the motion pattern of the target video. We follow its setting by adopting a LoRA rank of $r=32$. For optimization steps, we employ sufficient steps of $t=1000$, selecting the best results that balance the fidelity of the motion pattern and reference. The appearance transfer method~\cite{alaluf2024cross} employs Cross-image Attention~(CiA) for semantic matching between two images. It applies a contrast value $\beta$ to the attention map for more accurate transfer, alongside a guidance value $\alpha$ to diverge from the original appearance. We follow the setting with $\alpha=3.5$. We observed that the original setting $\beta=1.67$ may introduce significant structural changes. Thus, we traverse $\beta$ across $[1, 1.33, 1.67]$ and select the best, referred to as CiA~($\beta^*$). 
For our method, we set $k=1$ at Zero-Stage for all experiments as the default choice. We perform fine-tuning at a resolution of 512 and an optimization step $t=400$ by default, utilizing the default LoRA configuration from EasyControl~\cite{zhang2025easycontrol}, with $r=128$, $\alpha=128$.

\subsection{Comparison with Baselines}

\subsubsection{Qualitative Results}
Figure~\ref{fig:baselines} showcases a challenging scenario characterized by substantial object motion, dynamic background changes, and a complex color layout in the reference. AnyV2V shows considerable degradation in imaging quality when tackling such complex motion. I2VEdit adequately fits the motion within 600 steps for this case; however, at this point, the appearance of the reference can no longer effectively control the subsequent frames. This suggests that for videos with complex motion significantly deviating from the pretrained I2V domain, achieving a balance between motion fitting and reference control in I2V is challenging. 
The basic CiA method, which applies an optical contrast $\beta^*$ to attention maps for accurate transfer, alters the structure of the target frame. While this may be acceptable for appearance transfer tasks involving two different objects without a need for strict structure preservation, it is unsuitable for video editing, where maintaining structural consistency of the target frame is crucial. Furthermore, without incorporating more accurate correspondence mechanisms such as DIFT, there can be inaccurate matches leading to missing color on the car's body. Our Zero-Stage approach, augmented with DIFT correspondence guidance, ensures structure preservation and consistent appearance transfer of the car, while the Hero-Stage can further mitigate degradation. 

Figure~\ref{fig:ablation_mode} left show results in Blender-Color-Edit dataset. With their simpler motion, I2VEdit can strike a balance between motion fitting and reference control, achieving visually acceptable results at an intuitive level. However, imaging quality degradation still occurs~(zoom in on the red box).

We also compare our method with recent commercial services, Kling\footnote{https://app.klingai.com/global/}, with its online service of reference-based video editing. It supports mask tracking, which requires the user to click and select the region of the object to be edited and track it throughout the video. As shown in Figure~\ref{fig:ablation_swap_kling}, the commercial models, with their superior temporal modeling, achieve better appearance consistency compared to I2VEdit with open-source lightweight I2V model. However, in terms of appearance details, it still falls short of what our method can achieve. Additionally, their motion fidelity to the target video is not guaranteed, and the imaging quality also deteriorates.

\subsubsection{Quantitative Results}

In Table~\ref{tab:baselines}, we evaluate the methods quantitatively across three sub-tasks. For the General-Edit task, our method demonstrates superior performance in Motion Smoothness. CiA, with the optimally-searched attention contrast $\beta^*$, exhibits the highest Subject Consistency, but sacrifices the preservation of the target structure, resulting in the lowest PSNR on the anchor frame. As illustrated in Figure~\ref{fig:baselines}, with a changing background, it consistently utilizes the background from the reference frame rather than preserving the original background. This leads to the highest Subject Consistency score, yet obscures issues of appearance inconsistency and structural deformation of the car. Multiple indirect metrics need to be evaluated concurrently. Our method attains the best or second-best performance across all metrics. For the Colorization and Blender-Color-Edit tasks, which allow for strict evaluation with ground truth, our method clearly achieves the highest scores.

\subsection{Ablation Studies}
\begin{figure*}[h]
  \centering
  \includegraphics[width=0.95\linewidth]{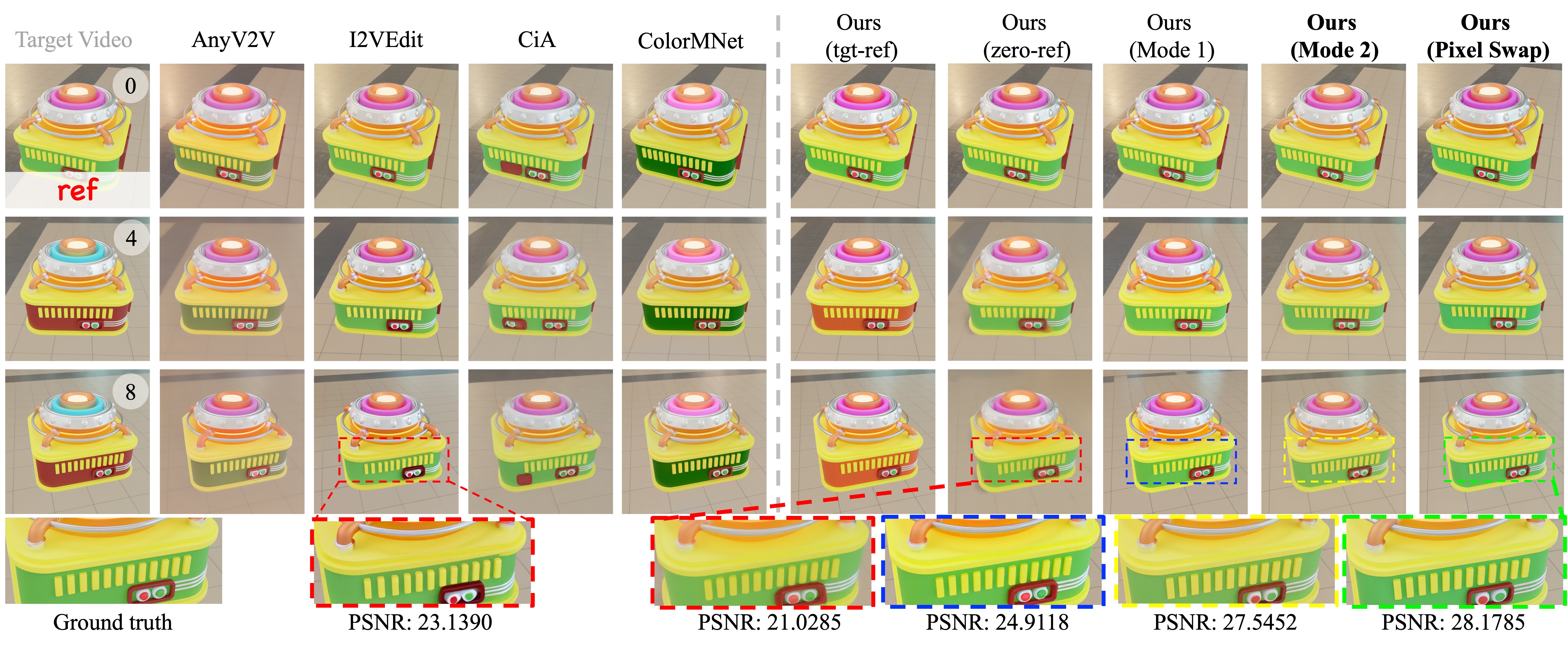}
  \caption{Qualitative results on Blender-Color-Edit dataset. \textbf{Left}: comparison with baselines. ColorMNet is proposed for colorization task. We convert target frames into grayscale for comparison, while it cannot transfer light green to dark region. \textbf{Right}: Ablations on training pairs. Mode 2 with Zero-Stage or Pixel Swap  initialization achieve higher PSNR than other pairs.}
  \label{fig:ablation_mode}
\end{figure*}
\subsubsection{Combinations of Training Data Pairs}

We have four paired datasets, as presented in Table~\ref{tab:data_pairs}. We also investigate the feasibility of directly learning the transfer from an anchor frame to a reference frame and generalizing this to subsequent target frames, referred to as tgt-ref. We conduct ablation studies using our constructed Blender-Color-Edit dataset to better discern the performance of each method. For each scenario, we perform training three times and compute the average to mitigate the effects of randomness in LoRA fine-tuning.



\begin{table}[t]
    \centering
    \resizebox{0.48\textwidth}{!}{\begin{tabular}{c|ccc|ccc}
    \toprule
      & \multicolumn{3}{c}{$t=600~(400 \text{ for Mode 2})$} & \multicolumn{3}{c}{$t=1000$}\\
      \midrule
     & avg & anc & $\text{tgt}_{(\Delta)}$ & avg & anc & $\text{tgt}_{(\Delta)}$ \\
    \midrule
    tgt-ref  & 19.38 & 22.29 & $17.52_{(-4.77)}$ & 22.39 & 27.09 & $18.62_{(-8.47)}$ \\
    zero-ref        & 24.09  & 29.57 & $20.95_{(-8.62)}$    & 24.59 & 31.80 & $20.92_{(-10.88)}$ \\
    \midrule
    Mode 1   & 24.54  & 28.79 & $22.08_{(-6.71)}$    & 25.77 & 30.92 & $22.91_{(-8.01)}$ \\
    Mode 2   & \textbf{26.76}  & 28.91 & $25.51_{(-\textbf{3.40})}$    & \textbf{27.38} & 29.97 & $25.65_{(\textbf{-4.32})}$ \\
    \bottomrule
    \end{tabular}}
    \caption{Ablation of training pairs on Blender-Color-Edit dataset: PSNR$(\uparrow)$ of avg: average across all frames; anc: anchor frame; tgt: last target frame; $\Delta$ represents difference between tgt and anc. For 1 condition~(tgt-ref, zero-ref, Mode 1), we evaluate at $t=600$, for 2 conditions~(Mode 2), we evaluate at $t=400$, to ensure the same optimization time consumption.}
    \label{tab:ablation_AB}
\end{table}

As shown in Table~\ref{tab:ablation_AB}, when directly fine-tuning the model to transfer the original anchor frame to the reference (tgt-ref) with sufficient $t=1000$ optimization steps, although the PSNR on the anchor frame reaches 27.09, the last target frame remains at 18.62, indicating a complete lack of generalization. As shown on the left side of Figure~\ref{fig:ablation_mode}, the anchor frame in the first row fits perfectly; however, in the subsequent frames (rows 2 and 3), there is noticeable leakage of the target appearance. 

With Zero-Stage as initialization, zero-ref achieves a higher PSNR across all target frames on average. However, the generalization ability is limited; even with more optimization steps, the gap between the anchor frame and the last target frame remains substantial~($\Delta=-10.88$). We then further introduce an auxiliary target frame. With Mode 1, this gap is reduced to 8.01. With Mode 2, the gap is further narrowed to 4.32, reaching the highest average PSNR of 27.38. Moreover, Mode 2 demonstrates notably faster convergence. Mode 2 with two conditions consumes 1.5 times the training time per step compared to the others with one condition. For fair comparison, we evaluate Mode 2 at $t=400$ while the others at $t=600$. Under the same optimization time, Mode 2 achieves the highest average PSNR and the smallest gap between the anchor frame and the last target frame.

\subsubsection{Zero-Stage Initialization}

\begin{figure*}[h]
  \centering
  \includegraphics[width=\linewidth]{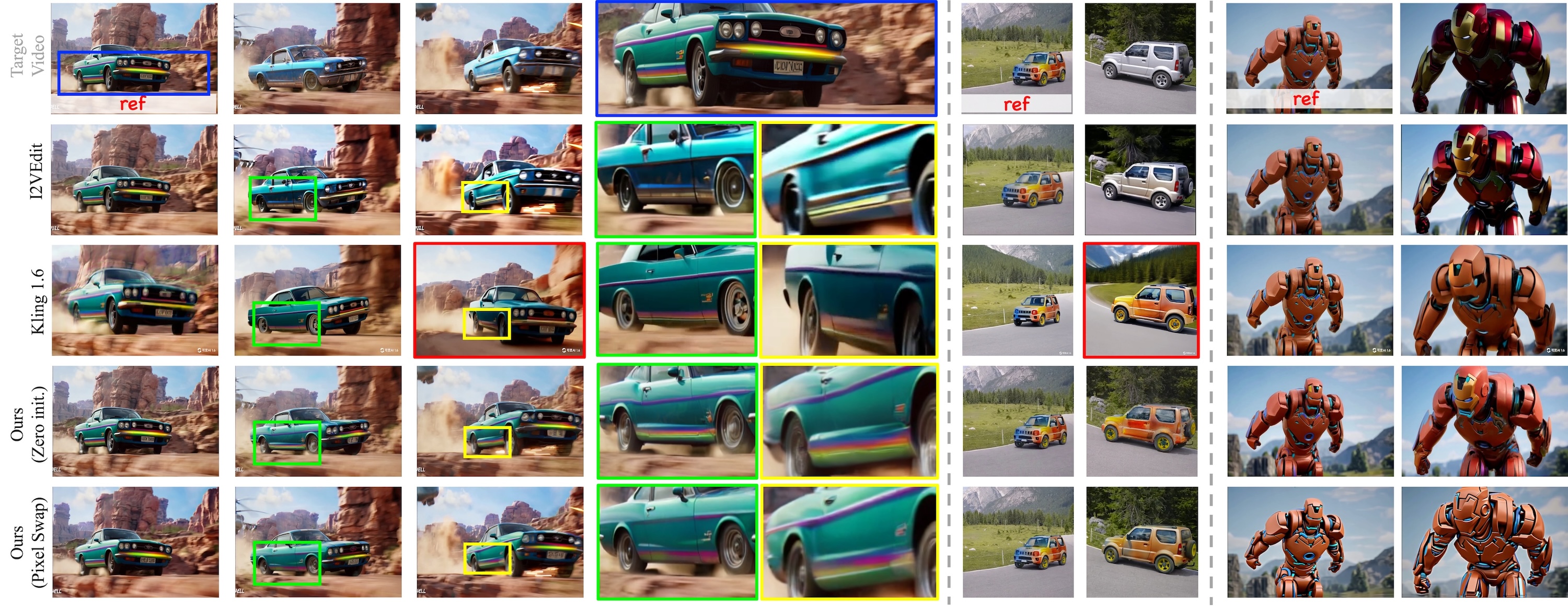}
  \caption{Qualitative results on General-Edit dataset. \textbf{Left}: Pixel swap initialization fails to provide sufficient information, resulting in some detailed appearances in the reference being missed in the subsequent frames (zoomed-in green and yellow box). \textbf{Middle/Right}: Pixel swap causes the fine-tuned model to become overly reliant on the target condition, hindering its ability to transfer the reference texture or style to subsequent frames. Kling online service of reference-based editing cannot strictly preserve target video motion~(red box).}
  \label{fig:ablation_swap_kling}
\end{figure*}


We emphasize the importance of Zero-Stage initialization. As illustrated in Table~\ref{tab:zero_init}, during the early optimization steps at $t=200$, our Zero-Stage initialization shows a faster increase in PSNR compared to pixel swap initialization and stabilizes at a slightly higher average PSNR by $t=400$. There is no obvious advantage visible at the visual level. Especially, when we analysis each case, we find some cases, pixel swap initialization achieve a higher PSNR score~(Figure~\ref{fig:ablation_mode}, last two columns). This is because, in the Blender-Color-Edit task, the color transformation involves a complete replacement of colors rather than complex color editing. Even with a pixel swap color layout lacking sufficient details, the model is still capable of blending the color layout into the target frame effectively.

For the General-Edit task with more complicated editing, the advantage of Zero-Stage initialization is more pronounced. As shown in Figure~\ref{fig:ablation_swap_kling}, pixel swap initialization fails to provide sufficient appearance information on details~(zoom in on the green and yellow box) as our Zero-Stage initialization. Also, in cases involving texture or style changes, pixel swap causes the fine-tuned model to overly depend on the target condition. As a result, it struggles to transfer the reference texture or style to subsequent frames.

\begin{table}[t]
    \centering
    \resizebox{0.35\textwidth}{!}
    {\begin{tabular}{c|cc}
    \toprule
    Setting & $t=200$ & $t=400$    \\
     \midrule
    {Pixel Swap}    & 22.4982 & 26.1908 \\
    {Zero-Stage Initialization}    & \textbf{23.7836} & \textbf{26.7640} \\
    
    \bottomrule
    \end{tabular}
   }
    \caption{Ablation on Zero-Stage initialization on Blender-Color-Edit dataset, which leads to a faster convergence in Hero-Stage and stabilize at a higher PSNR.}
    \label{tab:zero_init}
\end{table}






\section{Conclusion}

In this paper, we introduce \emph{Zero-to-Hero} for reference-based video appearance editing. By leveraging accurate correspondence to guide Cross-image Attention between reference and target frames, our Zero-Stage ensures fine-grained appearance transfer for videos with large motion. To address the compounded degradation caused by attention intervention, our Hero-Stage learns a conditional generative model based on training pairs from the anchor frame. This approach generalizes effectively across videos and ensures consistent, holistic restoration of the entire video. Experimental results demonstrate the effectiveness of our method, outperforming baselines in terms of high reference fidelity and video temporal consistency.

\section*{Acknowledgment}
This work is supported by Key Scientific Research Base for Digital Conservation of Cave Temples (Zhejiang University), State Administration for Cultural Heritage, and Alibaba Research Intern Program. Work done during T. Su's internship at Alibaba Cloud Computing. Correspondence to: C. Wang and D. Lu.


\begin{thebibliography}{47}
  \providecommand{\natexlab}[1]{#1}
  \providecommand{\url}[1]{\texttt{#1}}
  \expandafter\ifx\csname urlstyle\endcsname\relax
    \providecommand{\doi}[1]{doi: #1}\else
    \providecommand{\doi}{doi: \begingroup \urlstyle{rm}\Url}\fi
  
  \bibitem[Ahn et~al.(2024)Ahn, Cho, Min, Jang, Kim, Kim, Park, Jin, and Kim]{ahn2024self}
  Donghoon Ahn, Hyoungwon Cho, Jaewon Min, Wooseok Jang, Jungwoo Kim, SeonHwa Kim, Hyun~Hee Park, Kyong~Hwan Jin, and Seungryong Kim.
  \newblock Self-rectifying diffusion sampling with perturbed-attention guidance.
  \newblock In \emph{European Conference on Computer Vision}, pages 1--17. Springer, 2024.
  
  \bibitem[Alaluf et~al.(2024)Alaluf, Garibi, Patashnik, Averbuch-Elor, and Cohen-Or]{alaluf2024cross}
  Yuval Alaluf, Daniel Garibi, Or Patashnik, Hadar Averbuch-Elor, and Daniel Cohen-Or.
  \newblock Cross-image attention for zero-shot appearance transfer.
  \newblock In \emph{ACM SIGGRAPH 2024 Conference Papers}, pages 1--12, 2024.
  
  \bibitem[black-forest labs(2023)]{flux}
  black-forest labs.
  \newblock Flux.
  \newblock [Online] \url{https://github.com/black-forest-labs/flux}, 2023.
  
  \bibitem[Blattmann et~al.(2023)Blattmann, Dockhorn, Kulal, Mendelevitch, Kilian, Lorenz, Levi, English, Voleti, Letts, et~al.]{SVD}
  Andreas Blattmann, Tim Dockhorn, Sumith Kulal, Daniel Mendelevitch, Maciej Kilian, Dominik Lorenz, Yam Levi, Zion English, Vikram Voleti, Adam Letts, et~al.
  \newblock Stable video diffusion: Scaling latent video diffusion models to large datasets.
  \newblock \emph{arXiv preprint arXiv:2311.15127}, 2023.
  
  \bibitem[Chen et~al.(2023)Chen, Yu, Ge, Yao, Xie, Wu, Wang, Kwok, Luo, Lu, et~al.]{chen2023pixart}
  Junsong Chen, Jincheng Yu, Chongjian Ge, Lewei Yao, Enze Xie, Yue Wu, Zhongdao Wang, James Kwok, Ping Luo, Huchuan Lu, et~al.
  \newblock Pixart-alpha: Fast training of diffusion transformer for photorealistic text-to-image synthesis.
  \newblock \emph{arXiv preprint arXiv:2310.00426}, 2023.
  
  \bibitem[Chung et~al.(2024)Chung, Hyun, and Heo]{chung2024style}
  Jiwoo Chung, Sangeek Hyun, and Jae-Pil Heo.
  \newblock Style injection in diffusion: A training-free approach for adapting large-scale diffusion models for style transfer.
  \newblock In \emph{Proceedings of the IEEE/CVF Conference on Computer Vision and Pattern Recognition}, pages 8795--8805, 2024.
  
  \bibitem[Cong et~al.(2023)Cong, Xu, Simon, Chen, Ren, Xie, Perez-Rua, Rosenhahn, Xiang, and He]{cong2023flatten}
  Yuren Cong, Mengmeng Xu, Christian Simon, Shoufa Chen, Jiawei Ren, Yanping Xie, Juan-Manuel Perez-Rua, Bodo Rosenhahn, Tao Xiang, and Sen He.
  \newblock Flatten: optical flow-guided attention for consistent text-to-video editing.
  \newblock \emph{arXiv preprint arXiv:2310.05922}, 2023.
  
  \bibitem[Epstein et~al.(2023)Epstein, Jabri, Poole, Efros, and Holynski]{epstein2023diffusion}
  Dave Epstein, Allan Jabri, Ben Poole, Alexei Efros, and Aleksander Holynski.
  \newblock Diffusion self-guidance for controllable image generation.
  \newblock \emph{Advances in Neural Information Processing Systems}, 36:\penalty0 16222--16239, 2023.
  
  \bibitem[Feng et~al.(2024)Feng, Weng, Wang, Yuan, Bao, Luo, Chen, and Guo]{feng2024ccedit}
  Ruoyu Feng, Wenming Weng, Yanhui Wang, Yuhui Yuan, Jianmin Bao, Chong Luo, Zhibo Chen, and Baining Guo.
  \newblock Ccedit: Creative and controllable video editing via diffusion models.
  \newblock In \emph{Proceedings of the IEEE/CVF Conference on Computer Vision and Pattern Recognition}, pages 6712--6722, 2024.
  
  \bibitem[Geyer et~al.(2023)Geyer, Bar-Tal, Bagon, and Dekel]{geyer2023tokenflow}
  Michal Geyer, Omer Bar-Tal, Shai Bagon, and Tali Dekel.
  \newblock Tokenflow: Consistent diffusion features for consistent video editing.
  \newblock \emph{arXiv preprint arXiv:2307.10373}, 2023.
  
  \bibitem[Hertz et~al.(2022)Hertz, Mokady, Tenenbaum, Aberman, Pritch, and Cohen-Or]{hertz2022prompt}
  Amir Hertz, Ron Mokady, Jay Tenenbaum, Kfir Aberman, Yael Pritch, and Daniel Cohen-Or.
  \newblock Prompt-to-prompt image editing with cross attention control.
  \newblock \emph{arXiv preprint arXiv:2208.01626}, 2022.
  
  \bibitem[Huang et~al.(2024)Huang, He, Yu, Zhang, Si, Jiang, Zhang, Wu, Jin, Chanpaisit, et~al.]{huang2024vbench}
  Ziqi Huang, Yinan He, Jiashuo Yu, Fan Zhang, Chenyang Si, Yuming Jiang, Yuanhan Zhang, Tianxing Wu, Qingyang Jin, Nattapol Chanpaisit, et~al.
  \newblock Vbench: Comprehensive benchmark suite for video generative models.
  \newblock In \emph{Proceedings of the IEEE/CVF Conference on Computer Vision and Pattern Recognition}, pages 21807--21818, 2024.
  
  \bibitem[Kara et~al.(2024)Kara, Kurtkaya, Yesiltepe, Rehg, and Yanardag]{kara2024rave}
  Ozgur Kara, Bariscan Kurtkaya, Hidir Yesiltepe, James~M Rehg, and Pinar Yanardag.
  \newblock Rave: Randomized noise shuffling for fast and consistent video editing with diffusion models.
  \newblock In \emph{Proceedings of the IEEE/CVF Conference on Computer Vision and Pattern Recognition}, pages 6507--6516, 2024.
  
  \bibitem[kohya ss(2023)]{blur-controlnet}
  kohya ss.
  \newblock Gaussian deblur controlnet.
  \newblock [Online] \url{https://huggingface.co/kohya-ss/controlnet-lllite}, 2023.
  
  \bibitem[Ku et~al.(2024)Ku, Wei, Ren, Yang, and Chen]{ku2024anyv2v}
  Max Ku, Cong Wei, Weiming Ren, Huan Yang, and Wenhu Chen.
  \newblock Anyv2v: A plug-and-play framework for any video-to-video editing tasks.
  \newblock \emph{arXiv preprint arXiv:2403.14468}, 2024.
  
  \bibitem[Liu et~al.(2025)Liu, Wang, Su, Ten, Huang, Guo, and Jia]{liu2025understanding}
  Bingyan Liu, Chengyu Wang, Tongtong Su, Huan Ten, Jun Huang, Kailing Guo, and Kui Jia.
  \newblock Understanding attention mechanism in video diffusion models.
  \newblock \emph{arXiv preprint arXiv:2504.12027}, 2025.
  
  \bibitem[Liu et~al.(2024)Liu, Li, Zhang, Lan, and Liu]{liu2024stablev2v}
  Chang Liu, Rui Li, Kaidong Zhang, Yunwei Lan, and Dong Liu.
  \newblock Stablev2v: Stablizing shape consistency in video-to-video editing.
  \newblock \emph{arXiv preprint arXiv:2411.11045}, 2024.
  
  \bibitem[lllyasviel(2023{\natexlab{a}})]{color-controlnet}
  lllyasviel.
  \newblock Colorization controlnet.
  \newblock [Online] \url{https://huggingface.co/lllyasviel/sd_control_collection/blob/main/ioclab_sd15_recolor.safetensors}, 2023{\natexlab{a}}.
  
  \bibitem[lllyasviel(2023{\natexlab{b}})]{tile-controlnet}
  lllyasviel.
  \newblock Tile controlnet.
  \newblock [Online] \url{https://huggingface.co/lllyasviel/ControlNet-v1-1/blob/main/control_v11f1e_sd15_tile.pth}, 2023{\natexlab{b}}.
  
  \bibitem[Luo et~al.(2024)Luo, Dunlap, Park, Holynski, and Darrell]{luo2024diffusion}
  Grace Luo, Lisa Dunlap, Dong~Huk Park, Aleksander Holynski, and Trevor Darrell.
  \newblock Diffusion hyperfeatures: Searching through time and space for semantic correspondence.
  \newblock \emph{Advances in Neural Information Processing Systems}, 36, 2024.
  
  \bibitem[Mou et~al.(2023)Mou, Wang, Song, Shan, and Zhang]{mou2023dragondiffusion}
  Chong Mou, Xintao Wang, Jiechong Song, Ying Shan, and Jian Zhang.
  \newblock Dragondiffusion: Enabling drag-style manipulation on diffusion models.
  \newblock \emph{arXiv preprint arXiv:2307.02421}, 2023.
  
  \bibitem[Mou et~al.(2024{\natexlab{a}})Mou, Wang, Song, Shan, and Zhang]{mou2024diffeditor}
  Chong Mou, Xintao Wang, Jiechong Song, Ying Shan, and Jian Zhang.
  \newblock Diffeditor: Boosting accuracy and flexibility on diffusion-based image editing.
  \newblock In \emph{Proceedings of the IEEE/CVF Conference on Computer Vision and Pattern Recognition}, pages 8488--8497, 2024{\natexlab{a}}.
  
  \bibitem[Mou et~al.(2024{\natexlab{b}})Mou, Wang, Xie, Wu, Zhang, Qi, and Shan]{mou2024t2i}
  Chong Mou, Xintao Wang, Liangbin Xie, Yanze Wu, Jian Zhang, Zhongang Qi, and Ying Shan.
  \newblock T2i-adapter: Learning adapters to dig out more controllable ability for text-to-image diffusion models.
  \newblock In \emph{Proceedings of the AAAI conference on artificial intelligence}, pages 4296--4304, 2024{\natexlab{b}}.
  
  \bibitem[Ouyang et~al.(2024)Ouyang, Dong, Yang, Si, and Pan]{ouyang2024i2vedit}
  Wenqi Ouyang, Yi Dong, Lei Yang, Jianlou Si, and Xingang Pan.
  \newblock I2vedit: First-frame-guided video editing via image-to-video diffusion models.
  \newblock In \emph{SIGGRAPH Asia 2024 Conference Papers}, pages 1--11, 2024.
  
  \bibitem[Park et~al.(2020)Park, Zhu, Wang, Lu, Shechtman, Efros, and Zhang]{park2020swapping}
  Taesung Park, Jun-Yan Zhu, Oliver Wang, Jingwan Lu, Eli Shechtman, Alexei Efros, and Richard Zhang.
  \newblock Swapping autoencoder for deep image manipulation.
  \newblock \emph{Advances in Neural Information Processing Systems}, 33:\penalty0 7198--7211, 2020.
  
  \bibitem[Peebles and Xie(2023)]{peebles2023scalable}
  William Peebles and Saining Xie.
  \newblock Scalable diffusion models with transformers.
  \newblock In \emph{Proceedings of the IEEE/CVF international conference on computer vision}, pages 4195--4205, 2023.
  
  \bibitem[Qi et~al.(2023)Qi, Cun, Zhang, Lei, Wang, Shan, and Chen]{qi2023fatezero}
  Chenyang Qi, Xiaodong Cun, Yong Zhang, Chenyang Lei, Xintao Wang, Ying Shan, and Qifeng Chen.
  \newblock Fatezero: Fusing attentions for zero-shot text-based video editing.
  \newblock In \emph{Proceedings of the IEEE/CVF International Conference on Computer Vision}, pages 15932--15942, 2023.
  
  \bibitem[Qin et~al.(2023)Qin, Zhang, Yu, Feng, Yang, Zhou, Wang, Niebles, Xiong, Savarese, et~al.]{qin2023unicontrol}
  Can Qin, Shu Zhang, Ning Yu, Yihao Feng, Xinyi Yang, Yingbo Zhou, Huan Wang, Juan~Carlos Niebles, Caiming Xiong, Silvio Savarese, et~al.
  \newblock Unicontrol: A unified diffusion model for controllable visual generation in the wild.
  \newblock \emph{arXiv preprint arXiv:2305.11147}, 2023.
  
  \bibitem[Rombach et~al.(2022)Rombach, Blattmann, Lorenz, Esser, and Ommer]{LDM}
  Robin Rombach, Andreas Blattmann, Dominik Lorenz, Patrick Esser, and Bj{\"o}rn Ommer.
  \newblock High-resolution image synthesis with latent diffusion models.
  \newblock In \emph{Proceedings of the IEEE/CVF Conference on Computer Vision and Pattern Recognition}, pages 10684--10695, 2022.
  
  \bibitem[Song et~al.(2020)Song, Meng, and Ermon]{song2020denoising}
  Jiaming Song, Chenlin Meng, and Stefano Ermon.
  \newblock Denoising diffusion implicit models.
  \newblock \emph{arXiv preprint arXiv:2010.02502}, 2020.
  
  \bibitem[Tan et~al.(2024)Tan, Liu, Yang, Xue, and Wang]{tan2024ominicontrol}
  Zhenxiong Tan, Songhua Liu, Xingyi Yang, Qiaochu Xue, and Xinchao Wang.
  \newblock Ominicontrol: Minimal and universal control for diffusion transformer.
  \newblock \emph{arXiv preprint arXiv:2411.15098}, 2024.
  
  \bibitem[Tan et~al.(2025)Tan, Xue, Yang, Liu, and Wang]{tan2025ominicontrol2}
  Zhenxiong Tan, Qiaochu Xue, Xingyi Yang, Songhua Liu, and Xinchao Wang.
  \newblock Ominicontrol2: Efficient conditioning for diffusion transformers.
  \newblock \emph{arXiv preprint arXiv:2503.08280}, 2025.
  
  \bibitem[Tang et~al.(2023)Tang, Jia, Wang, Phoo, and Hariharan]{tang2023emergent}
  Luming Tang, Menglin Jia, Qianqian Wang, Cheng~Perng Phoo, and Bharath Hariharan.
  \newblock Emergent correspondence from image diffusion.
  \newblock \emph{Advances in Neural Information Processing Systems}, 36:\penalty0 1363--1389, 2023.
  
  \bibitem[Tewel et~al.(2024)Tewel, Kaduri, Gal, Kasten, Wolf, Chechik, and Atzmon]{tewel2024training}
  Yoad Tewel, Omri Kaduri, Rinon Gal, Yoni Kasten, Lior Wolf, Gal Chechik, and Yuval Atzmon.
  \newblock Training-free consistent text-to-image generation.
  \newblock \emph{ACM Transactions on Graphics (TOG)}, 43\penalty0 (4):\penalty0 1--18, 2024.
  
  \bibitem[Tumanyan et~al.(2022)Tumanyan, Bar-Tal, Bagon, and Dekel]{tumanyan2022splicing}
  Narek Tumanyan, Omer Bar-Tal, Shai Bagon, and Tali Dekel.
  \newblock Splicing vit features for semantic appearance transfer.
  \newblock In \emph{Proceedings of the IEEE/CVF Conference on Computer Vision and Pattern Recognition}, pages 10748--10757, 2022.
  
  \bibitem[Tumanyan et~al.(2023)Tumanyan, Geyer, Bagon, and Dekel]{tumanyan2023plug}
  Narek Tumanyan, Michal Geyer, Shai Bagon, and Tali Dekel.
  \newblock Plug-and-play diffusion features for text-driven image-to-image translation.
  \newblock In \emph{Proceedings of the IEEE/CVF Conference on Computer Vision and Pattern Recognition}, pages 1921--1930, 2023.
  
  \bibitem[Wang et~al.(2024)Wang, Ma, Guo, Xiao, Huang, and Li]{wang2024cove}
  Jiangshan Wang, Yue Ma, Jiayi Guo, Yicheng Xiao, Gao Huang, and Xiu Li.
  \newblock Cove: Unleashing the diffusion feature correspondence for consistent video editing.
  \newblock \emph{arXiv preprint arXiv:2406.08850}, 2024.
  
  \bibitem[Xu et~al.(2022)Xu, Zhang, Cai, Rezatofighi, and Tao]{xu2022gmflow}
  Haofei Xu, Jing Zhang, Jianfei Cai, Hamid Rezatofighi, and Dacheng Tao.
  \newblock Gmflow: Learning optical flow via global matching.
  \newblock In \emph{Proceedings of the IEEE/CVF conference on computer vision and pattern recognition}, pages 8121--8130, 2022.
  
  \bibitem[Xu et~al.(2024)Xu, Wang, Xiao, Liu, and Chen]{xu2024freetuner}
  Youcan Xu, Zhen Wang, Jun Xiao, Wei Liu, and Long Chen.
  \newblock Freetuner: Any subject in any style with training-free diffusion.
  \newblock \emph{arXiv preprint arXiv:2405.14201}, 2024.
  
  \bibitem[Yang et~al.(2023)Yang, Zhou, Liu, and Loy]{yang2023rerender}
  Shuai Yang, Yifan Zhou, Ziwei Liu, and Chen~Change Loy.
  \newblock Rerender a video: Zero-shot text-guided video-to-video translation.
  \newblock In \emph{SIGGRAPH Asia 2023 Conference Papers}, pages 1--11, 2023.
  
  \bibitem[Yang et~al.(2024{\natexlab{a}})Yang, Zhou, Liu, and Loy]{yang2024fresco}
  Shuai Yang, Yifan Zhou, Ziwei Liu, and Chen~Change Loy.
  \newblock Fresco: Spatial-temporal correspondence for zero-shot video translation.
  \newblock In \emph{Proceedings of the IEEE/CVF Conference on Computer Vision and Pattern Recognition}, pages 8703--8712, 2024{\natexlab{a}}.
  
  \bibitem[Yang et~al.(2024{\natexlab{b}})Yang, Dong, Tang, and Pan]{yang2024colormnet}
  Yixin Yang, Jiangxin Dong, Jinhui Tang, and Jinshan Pan.
  \newblock Colormnet: A memory-based deep spatial-temporal feature propagation network for video colorization.
  \newblock In \emph{European Conference on Computer Vision}, pages 336--352. Springer, 2024{\natexlab{b}}.
  
  \bibitem[Yang et~al.(2024{\natexlab{c}})Yang, Pan, Peng, Du, Tao, and Tang]{yang2024bistnet}
  Yixin Yang, Jinshan Pan, Zhongzheng Peng, Xiaoyu Du, Zhulin Tao, and Jinhui Tang.
  \newblock Bistnet: Semantic image prior guided bidirectional temporal feature fusion for deep exemplar-based video colorization.
  \newblock \emph{IEEE Transactions on Pattern Analysis and Machine Intelligence}, 2024{\natexlab{c}}.
  
  \bibitem[Zhang et~al.(2024)Zhang, Herrmann, Hur, Polania~Cabrera, Jampani, Sun, and Yang]{zhang2024tale}
  Junyi Zhang, Charles Herrmann, Junhwa Hur, Luisa Polania~Cabrera, Varun Jampani, Deqing Sun, and Ming-Hsuan Yang.
  \newblock A tale of two features: Stable diffusion complements dino for zero-shot semantic correspondence.
  \newblock \emph{Advances in Neural Information Processing Systems}, 36, 2024.
  
  \bibitem[Zhang et~al.(2023{\natexlab{a}})Zhang, Rao, and Agrawala]{zhang2023adding}
  Lvmin Zhang, Anyi Rao, and Maneesh Agrawala.
  \newblock Adding conditional control to text-to-image diffusion models.
  \newblock In \emph{Proceedings of the IEEE/CVF international conference on computer vision}, pages 3836--3847, 2023{\natexlab{a}}.
  
  \bibitem[Zhang et~al.(2023{\natexlab{b}})Zhang, Wang, Zhang, Zhao, Yuan, Qin, Wang, Zhao, and Zhou]{zhang2023i2vgen}
  Shiwei Zhang, Jiayu Wang, Yingya Zhang, Kang Zhao, Hangjie Yuan, Zhiwu Qin, Xiang Wang, Deli Zhao, and Jingren Zhou.
  \newblock I2vgen-xl: High-quality image-to-video synthesis via cascaded diffusion models.
  \newblock \emph{arXiv preprint arXiv:2311.04145}, 2023{\natexlab{b}}.
  
  \bibitem[Zhang et~al.(2025)Zhang, Yuan, Song, Wang, and Liu]{zhang2025easycontrol}
  Yuxuan Zhang, Yirui Yuan, Yiren Song, Haofan Wang, and Jiaming Liu.
  \newblock Easycontrol: Adding efficient and flexible control for diffusion transformer.
  \newblock \emph{arXiv preprint arXiv:2503.07027}, 2025.
  
  \end{thebibliography}
\end{document}